# Analyzing Self-Driving Cars on Twitter

Rizwan Sadiq [1] and Mohsin Khan [2]

[1] Lecturer at COMSATS Institute of Information Technology, Abbottabad, Pakistan
`rizwansadiq@ciit.net.pk`

[2] International Islamic University, Islamabad, Pakistan
`mohsin.khan@iiu.edu.pk`

**Abstract:**

This paper studies users' perception regarding a controversial product, namely self-driving (autonomous) cars. To find people's opinion regarding this new technology, we used an annotated Twitter dataset, and extracted the topics in positive and negative tweets using an unsupervised, probabilistic model known as topic modeling. We later used the topics, as well as linguist and Twitter specific features to classify the sentiment of the tweets. Regarding the opinions, the result of our analysis shows that people are optimistic and excited about the future technology, but at the same time they find it dangerous and not reliable. For the classification task, we found Twitter specific features, such as hashtags as well as linguistic features such as emphatic words among top attributes in classifying the sentiment of the tweets.

## Introduction

Every day, new technologies and models in the area of Artificial Intelligence (AI) are introduced to the world. One of the recent controversial technologies is autonomous or self-driving cars. Tech companies such as Google and Tesla are competing with each other to create and attract more people to purchase their products. While we find these improvements in technologies very helpful, we believe that these companies first need to study the impact and consequences of the new technologies which are completely strange to people or society. As some researchers noted before, new technologies can be helpful if the society, urban design, and citizens of that society are capable of accepting and implementing new changes (Brynjolfsson and Mitchell, 2017).

On the other hand, users are sometimes positively biased toward accepting the new products because of 1) their personal experience and efficiency of the machines, and 2) fewer errors of machine compared to humans. Overestimating the reliability of the systems and being distant from the truth (about the algorithms, failures of the machine and etc.) by investors and business owners results in more trust on automated systems by the users. In their study, Lyons and Stokes (2012) compared the reliance of the users on human vs. automated systems and found that users tend to rely on the latter one especially in vulnerable situations.

New technologies such as chatbots, automatic navigation systems, autonomous cars, and even internet and web influence the structure of users lives. Obviously, they are all created to provide better services to the users. However, over-reliance on the systems as well as shortcomings in integrating users' expectations, in some cases, resulted in serious failures. For instance, due to unfamiliarity with the environment, two people were injured and killed because of the recommended route by a smart phone navigating system (Liberatore, 2016). This example highlights the importance of studying and analyzing people's opinion and experiences to minimize the risk for the end-users. With the increase in people's interest in using self-driving cars, there is a huge need for companies to find the users' viewpoints and integrate the new features in their future devices.

In this study we aim to fill this gap by exploring and understanding the opinion of online communities on social media. Moreover, we first extract users' opinion regarding the autonomous cars using an unsupervised learning method. We then study the sentiment of the tweets that people write and share on the topic of automated cars. In another part of this study, we implement a computational method to analyze the sentiment of the user-generated text on Twitter and predict the polarity of the tweets using various textual features. The result of this research would help the designers and tech companies to have a better understanding of citizens of the societies that are the recipients of their products.

The rest of the paper is organized as follows: the related work (section 2) discusses the prior work in the area of sentiment analysis. We then discuss the data collection and methodology, where the methodology section is divided to supervised and unsupervised learning. After that, we report the result of each model. The last section discusses the conclusion and future directions for improving the paper.

## Related work

**Sentiment Analysis**

In recent years, availability of huge online text opened various research domains in the area of natural language processing. Subjectivity or sentiment analysis is one of the classification technique that assigns a best fitting category as positive, negative and neutral to text documents (Shanahan at al., 2006). This set of categories is particularly appropriate for labeling most of the user-generated texts such as reviews, tweets, and blogs (Pang et al., 2008). There are two commonly used approached for analyzing the sentiment of texts, 1) the lexicon-based approach and 2) machine learning methods. Based on a survey, it was found that lexicon-based approaches are sometimes more effective compared to the other methods (Taboada et al., 2011). However, numbers of studies found the classification method more reliable and generalizable. Moreover, the accuracy of lexicon-based approaches sometimes closely depend on the context but using meaningful features, classifiers can achieve a better performance to predict the polarity of the texts. Sentiment analysis has been widely used in review mining to extract the opinion of people about a product. Online reviews are one of the important resources for both industry owners and customers to find the feedbacks or positive and negative opinions about a product. Therefore, various methods have been used to study the relationship between opinion and sentiment in this domain (Liu et al., 2012; Zhao et al, 2009, Hong et al., 2012)

In the past couple of years, analyzing microblogging data, namely tweets, attracted more research in the area of sentiment analysis. One unique feature of the microblogs is that they reflect people's behavior and emotion regarding the events in society. Therefore, this valuable source of information attracted many scholars to analyze tweets and their content deeper to find efficient automatic techniques and algorithms to estimate the polarity and understand opinions or even impact of events and products on individuals and citizens.

In their study, O'Connor and group (2010) studied supplementing the polling system using number of tweets. Moreover, they used polling numbers as the ground truth of the study, and leveraged sentiment word frequencies to label the tweet as positive or negative. They found that social media data can supplement traditional polling. In another study, Kouloumpis and group (2011) studied the usefulness of various linguistic and Twitter specific features to analyze the sentiment of the tweets. They found part of speech and linguistic features not very useful for their task. However, the twitter specific features, such as emoticons were extremely helpful. In addition, Agarwal and group (2011) analyzed the usage of new methods as "tree kernel and feature-based models" and found both more efficient compared to the unigram baseline. Rezapour and group (2017) studied the likability of the 2016 US presidential election and extracted the opinion of the Twitter users about both Republican and Democrat candidates in the New York's primary election. In their lexicon-based approach, they extracted and annotated the hashtags with their polarity and then added them to the polarity lexicon. Their study showed that the new hashtag-enhanced lexicon resulted in a better performance compared to other methods. They extracted the users' opinion regarding the candidates and found Trump and Sanders the most favorite candidates in republican and democratic parties. Moreover, some researchers tried to analyze the polarity of hashtags using the content of the tweets. In their paper, Tsur and Rappoport (2012) collected a dataset of four hundred millions of tweets. They then extracted several features such as sentiment, psychological, temporal and number of retweets. Using a regression model, the predicted the "exposure" of hashtags.

The reviewed literature above is just a very small sample of the study in this domain. In this paper, we will leverage a classification method and use the features that were found the most useful for the purpose of sentiment analysis in the reviewed literature.

# Data

The best source of data to get the individuals' opinion is social media platforms such as Twitter and Facebook. Due to some privacy restrictions extracting and using the Facebook data is challenging. The API does not allow the person to collect friend's data unless each person in the network issues an official consent. Although there are various questionable ways to get that data, we preferred to use Twitter as the input of our study since it makes it easier for researchers to extract, and share the data. For this study, we used an annotated twitter dataset on the topic of self- driving cars which is created and distributed by CrowdFlower[1]. The dataset consists of 7015 tweets labeled from 1 to 5, as 1 showing the most negative tweet and 5 showing the most positive ones. The tweets not related to the topic of self-driving cars are labeled as "non-relevant." While this category can be useful in some studies, we excluded them for the purpose of our research. After removing the non-relevant tweets, the final dataset consists of 6943 tweets. Table 1 shows the number of the tweets in each category.

|  | Sentiment | | | | | |
| --- | --- | --- | --- | --- | --- | --- |
|  | 5 | 4 | 3 | 2 | 1 | total |
| # tweets in the original dataset | 459 | 1444 | 4245 | 685 | 110 | 6943 |

# Method

**Unsupervised learning:**

**Topic Modeling:** One of the popular methods in the area of information extraction and text summarization is topic modeling (Blei, 2012; Blei, Ng, and Jordan, 2003). Topic modeling is a statistical model for extracting the latent semantic structure of a dataset. Using topic modeling, we can extract the gist of the documents and understand what the most salient themes in a body of text are. In this paper, we used Mallet a topic-modeling package that leverages Gibbs sampling to infer topics of the documents (McCallum, 2002). To find the general opinion of the people regarding the autonomous cars, we divided the data to two sets as positive and negative and ran the model on the datasets. Figures 1 and 2 show the positive and negative keywords extracted from the tweets.

**Supervised Learning:**

**Features**

To classify the sentiment of this dataset, we chose three different feature set: (1) Top Unigrams, (2) Top Linguistic Features, and (3) Meta-data.

*Top Unigrams:* we first added the top TF-IDF word to this set of features. This metric returns the most informative words in the corpus. To get these words, after removing the stop words and stemming the tokens, we extracted the top 100 salient unigrams from the data. We created the baseline feature set using these words. In addition, we added the top topic words to the word features if they were not among the top TF-IDF words.

---

[1] https://www.crowdflower.com/data-for-everyone/

*Linguistic Features:* to have a better understating of the nature of the tweets we considered, the number of verbs, nouns, adjective and adverbs in the dataset. We believe that the topic of self-driving cars is currently among the top controversial topics in the USA. Previously, some works found emphatic words helpful for predicting controversial topics (Addawood et al., 2017). Therefore, we extracted empathic words form the dataset and added them to linguistic features.

*Meta-Data:* tweets are accompanied by additional information such as Retweets, URLs, Hashtags, number of followers and followed. We extracted the number of presence of each of these information for each tweet and created the Meta-Data feature set. In addition, based on the literature in the domain of sentiment analysis, it was found that hashtags are among important attributes in tweets. One study found hashtags helpful in predicting the sentiment of tweets in political domain (Rezapour et al., 2017). To leverage the hashtags, we first extracted all of them from the dataset, tagged them with their polarity (if possible) and then counted the presence of positive, neutral, and negative hashtags in each tweet.

**Classifier**

To classify the tweets we used the Random Forest algorithm, one of the popular algorithms in the domain of NLP and machine learning. This algorithm leverages a modified tree learning model which selects a random set of features. We conducted 10 fold cross-validations, and used standard metrics as F1, Precision and Recall to assess the prediction. To implement the classifiers, we used Weka (Hall et al., 2009). Before running the algorithm, Information Gain was used to select the most beneficial attributes in our feature sets. We created a baseline using the unigram feature set in the beginning and then added the additional feature sets one by one on top of the baseline to find the best combination of features for classifying the sentiment on our self-driving dataset. The result of the prediction is discussed in the following section (Table 1).

## Result

**Topic Modeling**

After running the topic modeling over the positive and negative texts, we created two-word clouds to visualize the extracted opinion of the users on Twitter. Figure 1 shows the positive topics while Figure 2 highlights the negative ones in our corpus. As shown in the first figure, people used positive words such as "perfect, nice, cool, faster, excited, future lead" to express their opinion and feelings regarding the self-driving cars. It shows that the supportive people are excited about the future of car companies and their promised. In addition, they indicate emotional support of the individuals for these products. Words such as "night" and "vision" can indicate individuals' hope for using autonomous cars at nights with less vision. Moreover, the users believe that these products will "reshape" the future.

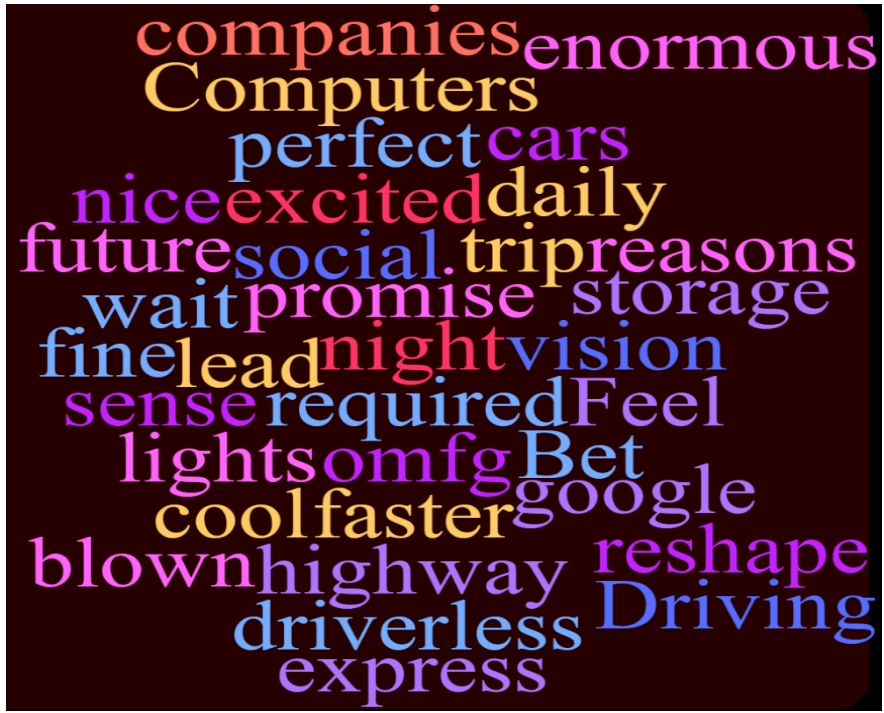

Figure 1: Top topics in the positive tweets

Regarding the negative dataset, as shown in the second figure, we can see that people find the cars "ridiculous, and difficult." They also indicated the problems with crowded highways and cities, crashes, and bikes, which shows their fear for incidents and reliability of these products. We also see words such as "compete, backseat, and talking" among the extracted topics which all show lack of support of this group for the autonomous cars.

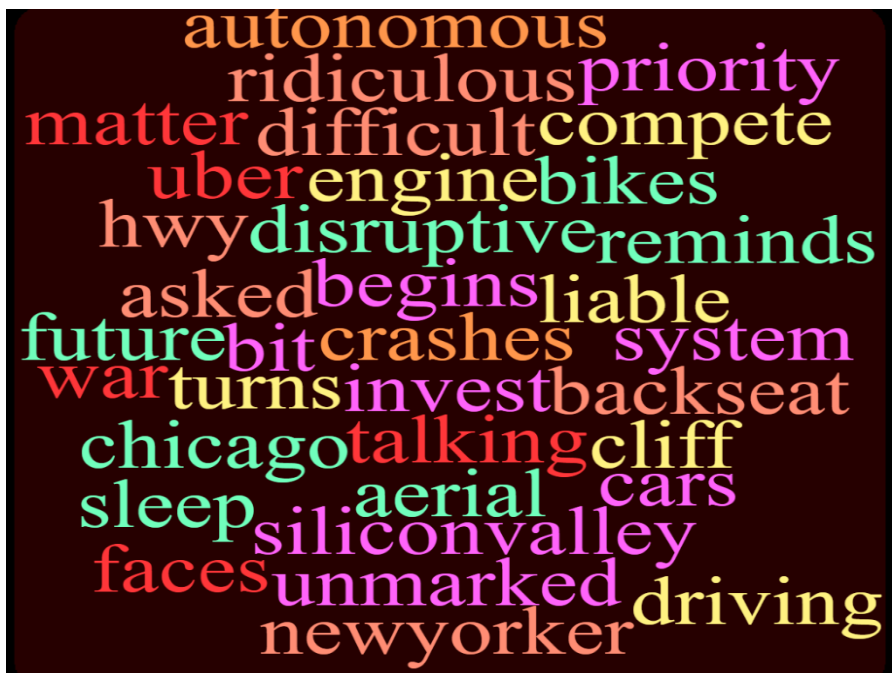

Figure 2: Top topics in the negative tweets

As the results above show, some people are excited to see the new technologies, while others are skeptical about them. One idea would be using aspect-based sentiment analysis to find the features that people were satisfied with. The observed topics in this section is intuitive and helpful to understand the perception of public, however, a more in-depth analysis would be beneficial to extract the socio-technical influence of the new technologies. The latter one is not the focus of this paper. Therefore, we will investigate the aspects in the future works.

**Classification**

As the result of our classifier shows (Table 1), using the baseline features achieved 55.6% accuracy. Adding the linguistic features on top of the baseline model, resulted in a 6% jump in accuracy. As found in prior studies, we also observed that the combination of linguistic features and Twitter specific features outperformed the unigram baseline (Kouloumpis, Wilson and Moore, 2011). The combination of these two resulted in 58.5% accuracy.

In addition, we used the combination of all these three features in the classifier and achieved the highest accuracy as 62.24%. This result show that adding twitter meta-data on top of the second feature combination (linguistic + unigrams) increased the accuracy by around 1%.

|  | Precision | Recall | F-Measure | Accuracy (%) |
|---|---|---|---|---|
| **Unigrams (Baseline)** | 0.494 | 0.556 | 0.517 | 55.61 |
| **Unigram + Linguistic** | 0.537 | 0.617 | 0.530 | 61.71 |
| **Linguistic + Meta-Data** | 0.462 | 0.585 | 0.490 | 58.50 |
| **Unigram + Meta-Data + Linguistic** | **0.562** | **0.622** | **0.528** | **62.24** |

**Table 1: Result of each feature set using Random Forest algorithms**

**Feature Analysis**

To have a better understanding about the classifiers, we extracted the top ten features (if applicable) of each combination of the feature sets as used in the classifier. Regarding the unigram, as show in Table 2, we found that words such as cool, awesome, drive google are among the top attributes. Interestingly, most of these words are the topic words extracted in the previous section (Figures 1 and 2). It shows that the topic words added to the top unigrams were among the efficient features in classifying the sentiment of the tweets.

Moreover, regarding the second feature set, attributes such as emphatics, length of tweets, adjective and adverbs as well as words such as awesome and cool are among the top features. It shows that, as we hypothesized, the linguistic features were helpful in the prediction. The top attributes in the combination of meta-data and linguistic features are Emphatics, length of tweets, adverb, adjective, verb, noun, and finally "selfdrivingcars". As mentioned in the method section, we extracted the hashtags and added them to the meta-data features. Having a hashtag among the top features show that the hashtags are helpful in analyzing sentiment of tweets, as found in previous studies (Rezapour et al., 2017). Finally, the top attributes in the combination of all features consist of the previously mentioned attributes as well as URLs.

The result of this section shows that classifying sentiment analysis is highly correlated with domain, context, and topic. Twitter consist of various embedded features that can assist in finding the polarity of the texts. In addition, the topic of the dataset can help in extracting meaningful features. For instance, as found previously, emphatic features are important in finding the controversial topics in twitter (Addawood et al., 2017). Since the topic of autonomous cars is among controversial topics in societies, we also found emphatic words among helpful features in our analysis.

|  | Top Attributes |
|---|---|
| **Unigrams (Baseline)** | Cool, awesome, wait, drive, car, don't problem, Google, people, driver |
| **Unigram + Linguistic** | Emphatics, cool, awesome, tweet_length, drive, wait, car, problem, adjective, adverb |
| **Linguistic + Meta-Data** | Emphatics, tweet_length, adverb, adjective, verb, noun, selfdrivingcars |
| **Unigram + Meta-Data + Linguistic** | URL, emphatics, cool, awesome, tweet_length, drive, adverb, car, adjective, don't |

**Table 2: Top attributes of each feature set using Information Gain**

## Conclusion and Future work

Autonomous cars are among the newest products of artificial intelligence. Many companies such as Tesla and Google are spending thousands of dollars to promote this technology. The concept of self-driving cars is still in debate and is regarded as a controversial topic in various societies. In this paper, we focused on extracting people's opinion regarding this new product and find if they like or dislike self-driving cars. Moreover, we used a tagged (positive and negative) user-generated texts (tweets) related to this topic. We first extracted the topics in positive and negative corpus and found that some people are optimist about the future of these products, and excited to see the new technologies. However, some others find self-driving cars ridiculous and dangerous. In the next step, to classify the sentiment of the tweets, we used different sets of features that we hypothesized to be useful for this task. We found that linguistic features, top topic words and the meta-data such as hashtags and tweet-length are the most useful features for classifying the sentiment of the tweets.

One major issue in the field of natural language processing is finding a reliable annotated dataset. In the future, we will focus on expanding the dataset to extract more user-generated texts from various domains such as reviews, and blogs. We believe that combining these domains will result in better understanding the users and citizens of a society who are the final recipient of the new products. We will also expand the feature sets for the concept of predicting the polarity of the input text. We are aware that deep learning algorithms are among the popular models in analyzing sentiment. One challenge in using neural nets are the lack of interpretability of the outcome. However, to boost the accuracy we plan to use these models in the future.

Another major challenge for big tech corporations is the processing of big data which is extracted from many tweets and online feedback from users. To overcome this issue, we can use MapReduce framework (Dean and Ghemawat, 2017) for processing the big data combined with efficient near-data load balancing algorithms such as the priority algorithm (Xie and Lu, 2015), the Weighted-Workload algorithm (Xie et al., 2016, Yekkehkhany, 2017), the GB-PANDAS algorithm (Yekkehkhany et al., 2017), and the MaxWeight algorithm (Stolyar, 2004, Wang et al., 2016). Note that the default scheduler for Hadoop (White, 2012) is the first-come-first-served scheduler which is not optimal and can lead to tremendous waste of resource. It is shown by (Wang et al., 2016) that by using the mentioned algorithms, not only the response time for processing the big data decreases, but it can also lead to saving of energy in data centers. For heuristic algorithms for near-data scheduling, refer to (Chen et al., 2012, Lin et al., 2013, Tan et al., 2013, He et al., 2011, Ibrahim et al., 2012, Isard et al., 2009, Jin et al., 2011).